\documentclass[11pt,a4paper]{article}
\usepackage[hyperref]{naaclhlt2018}
\usepackage{times}
\usepackage{latexsym}
\usepackage{multirow}
\usepackage{graphicx}
 \graphicspath{{figures/}}
\usepackage{url}
\usepackage{covington}
\usepackage{enumerate}
\usepackage{enumitem}
 \setenumerate{noitemsep,topsep=0pt,parsep=0pt,partopsep=0pt,leftmargin=0.3cm}
\usepackage{amsfonts, amsmath, amsthm, amssymb}
\usepackage{multirow}
\usepackage{graphicx}
\usepackage{fixltx2e}
\usepackage[english]{babel}
\usepackage[utf8x]{inputenc}
\usepackage[font=small]{caption}
\usepackage{subcaption}
\usepackage{tikz-qtree}
\usepackage{tikz}
\usetikzlibrary{calc,matrix,decorations.markings,decorations.pathreplacing,external}
\usepackage{algorithm}
\usepackage[noend]{algpseudocode}
\makeatletter
\def\BState{\State\hskip-\ALG@thistlm}
\makeatother
\usepackage{newfloat}
 \DeclareFloatingEnvironment[name={Example}]{row}

\newcommand{\pt}{1.5pt}
\newcommand{\globalcaptionsetup}{-10pt}
\aclfinalcopy
\def\aclpaperid{966}

\title{Diachronic Usage Relatedness (DURel):\\
A Framework for the Annotation of Lexical Semantic Change}

\author{$^{*}$Dominik Schlechtweg, $^{*}$Sabine Schulte im Walde, $^{\dag}$Stefanie Eckmann\\
$^{*}$Institute for Natural Language Processing, University of Stuttgart, Germany \\
$^{\dag}$Historical and Indo-European Linguistics, LMU Munich, Germany \\
{\tt \small dominik.schlechtweg@ims.uni-stuttgart.de}, {\tt \small schulte@ims.uni-stuttgart.de}, \\{\tt \small stefanie.eckmann@campus.lmu.de}
 }

\date{}

\begin{document}
\maketitle

\begin{abstract}
We propose a framework that extends synchronic polysemy annotation to diachronic changes in lexical meaning, to counteract the lack of resources for evaluating computational models of lexical semantic change. Our framework exploits an intuitive notion of semantic relatedness, and distinguishes between innovative and reductive meaning changes with high inter-annotator agreement. The resulting test set for German comprises ratings from five annotators for the relatedness of 1,320 use pairs across 22 target words.
\end{abstract}

\section{Introduction}

We see an increasing interest in the automatic detection of semantic change in computational linguistics \citep[i.a.]{Hamilton16,Frermann:2016,schlechtweg-EtAl:2017:CoNLL}, motivated by expected performance improvements of practical NLP applications, or theoretical interest in language or cultural change. However, a major obstacle in the computational modeling of semantic change is evaluation \citep{Lau12p591,Cook14p1624,Frermann:2016,dubossarsky2017}. Most importantly, there is no reliable test set of semantic change for any language that goes beyond a small set of hand-selected targets. We counteract this lack of resources by extending a framework of synchronic polysemy annotation to the annotation of Diachronic Usage Relatedness (DURel). DURel has a strong theoretical basis and at the same time makes use of established synchronic procedures that rely on the intuitive notion of semantic relatedness. The annotations distinguish between innovative and reductive meaning change with high inter-annotator agreement. DURel is language-independent and thus applicable across languages; this paper introduces the first test set of lexical semantic change for German.

\section{Related Work}
\label{sec:relatedwork}
A large number of studies has been performed on synchronic word sense annotation (see \citealp{IdePustejovsky2017} for an overview). Within this set, our paper is most related to work focusing on graded polysemy annotation. Most prominently, \citet[][]{SoaresdaSilva1992} is interested in the question of whether the theoretical distinction between polysemy and homonymy can be experimentally verified; \citet{Brown2008} wants to know how fine-grained word senses are, and \citet{Erk09investigationson,Erk13} examine whether we should adopt a graded notion of word meaning.

In contrast, there is little work on annotation with a focus on semantic change, despite the growing interest and modeling efforts in the field of semantic change detection. \citet[][]{Lau12p591} and \citet[][]{Cook14p1624} aim at verifying the semantic developments of their targets by a quasi-annotation procedure of dictionary entries, however without reporting inter-annotator agreement or other measures of reliability. \citet[][]{Gulordava11} ask annotators for their intuitions about changes but without direct relation to language data. \citet[][]{Bamman11p1} exploit aligned translated texts as source of word senses and conduct a very limited annotation study on Latin texts from different time periods. \citet{schlechtweg-EtAl:2017:CoNLL} propose a small-scale annotation of metaphoric change, but altogether there is no standard test set across languages that goes beyond a few hand-selected examples.

\section{Lexical Semantic Change}
\label{sec:change}

It is well-known that lexical semantic change and polysemy are tightly connected. For example, \citet{Blank97XVI} develops an elaborate theory where polysemy is the synchronic, observable result of lexical semantic change. He distinguishes two main types of lexical semantic change:

\begin{itemize}
\item \textbf{innovative meaning change}: emergence of a full-fledged additional meaning of a word; old and new meaning are related by polysemy
\item \textbf{reductive meaning change}: loss of a full-fledged meaning of a word
\end{itemize}
An example of innovative meaning change is the emergence of polysemy in the German word \textit{Donnerwetter} around 1800 \citep{Paul02XXI}. Before $\approx$1800 \textit{Donnerwetter} was only used in the meaning of `thunderstorm'. After 1800 we still observe this meaning, and in addition we find a new, clearly distinguished meaning as a swear word `Man alive!'.
An example of reductive meaning change is the German word \textit{Zufall}. It had two meanings $\approx$1850, `seizure' and `coincidence' \citep{Osman1971}. After 1850, the word occurs less and less often in the former meaning, until it is exclusively used in the meaning of `coincidence'. \textit{Zufall} lost the meaning `seizure'.

\subsection{Semantic Proximity} 
Based on Prototype Theory \citep{Rosch1975}, \citeauthor{Blank97XVI} develops criteria to decide whether word uses are related by polysemy. He defines a continuum of \textit{semantic proximity} with polysemy located between identity and homonymy, as depicted in Table \ref{tab:blank}. 

\begin{table}[htp]
\begin{center}
\tabcolsep=0.11cm
\scalebox{1}{
\begin{tabular}{ll}
\multirow{4}{*}{$\Bigg\uparrow$}&Identity\\
&Context Variance\\
&Polysemy\\
&Homonymy
\end{tabular}}
\caption{Continuum of semantic proximity \citep[cf.][p.~418]{Blank97XVI}.}
\label{tab:blank}
\end{center}
\end{table}
While it is difficult to directly apply these criteria to practical annotation tasks, we exploit the scale of semantic proximity indirectly, as previously done by synchronic research on polysemy applying similar scales \citep{SoaresdaSilva1992,Brown2008,Erk13}. Especially \citeauthor{Erk13}'s in-depth study validates an annotation framework relying on a scale of semantic proximity, revealing high inter-annotator agreement and strong correlation with traditional single-sense annotation as well as annotation of multiple lexical paraphrases. For our study, we decided to adopt a relatedness scale similar to \citeauthor{Brown2008}'s, shown in Table \ref{tab:scale2}.

\captionsetup{belowskip=-50pt}
\begin{table}[htp]
\begin{center}
\tabcolsep=0.11cm
\scalebox{1}{
\begin{tabular}{ll}
\multirow{4}{*}{$\Bigg\uparrow$} &4: Identical\\
 &3: Closely Related\\
 &2: Distantly Related\\
 &1: Unrelated\\
 & \\
 &0: Cannot decide
\end{tabular}}
\caption{Our 4-point scale of relatedness derived from \citet{Brown2008}.}
\label{tab:scale2}
\end{center}
\end{table}
\captionsetup{belowskip=\globalcaptionsetup}

\subsection{Diachronic Usage Relatedness (DURel)}

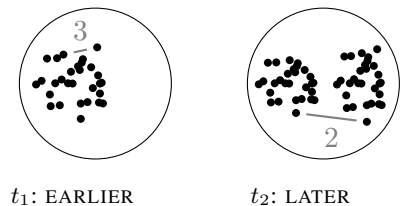
\begin{figure}[htbp]
\newcommand{\xslant}{1.2}
\newcommand{\yslant}{1.2}
\begin{center}
\def\firstcircle{(0,0) circle (1cm)}
\def\secondcircle{(0:3cm) circle (1cm)}
\begin{tikzpicture}
    \begin{scope}[shift={(3cm,-5cm)}, fill opacity=0.55]
    \draw \firstcircle;
    \draw \secondcircle;
        \end{scope}
        \begin{scope}[shift={(2.7cm,-5cm)}]
        
            \fill (0,-1.5) node {\small $t_1$: \textsc{earlier}};
            \fill (3,-1.5) node {\small $t_2$: \textsc{later}};
        
		\fill (\xslant*0,0*\yslant) circle (\pt) node ('A') {};
		
		\fill (\xslant*0.079,0.15*\yslant) circle (\pt) node ('B') {};
       		\fill (\xslant*-0.12,0.04449*\yslant) circle (\pt) node ('C') {};
		\fill (\xslant*0.06,0.19*\yslant) circle (\pt) node ('D') {};
		\fill (\xslant*0.3,-0.11*\yslant) circle (\pt) node ('E') {};
		
		\fill (\xslant*0.204,0.18*\yslant) circle (\pt) node ('F') {};
       		\fill (\xslant*-0.05,0.003*\yslant) circle (\pt) node ('G') {};
		\fill (\xslant*0.1,0.23*\yslant) circle (\pt) node ('H') {};
		\fill (\xslant*0.09,-0.221*\yslant) circle (\pt) node ('I') {};

		\fill (\xslant*0.3,0*\yslant) circle (\pt) node ('J') {};
       		\fill (\xslant*-0.34,0.03*\yslant) circle (\pt) node ('K') {};
		\fill (\xslant*-0.1,0.32*\yslant) circle (\pt) node ('L') {};
		\fill (\xslant*0.09,-0.39*\yslant) circle (\pt) node ('M') {};

		\fill (\xslant*0.28,0.09*\yslant) circle (\pt) node ('N') {};
       		\fill (\xslant*-0.0178,0.13*\yslant) circle (\pt) node ('O') {};
		\fill (\xslant*0.113,0.269*\yslant) circle (\pt) node ('P') {};
		\fill (\xslant*0.164,-0.205*\yslant) circle (\pt) node ('Q') {};
		
		\fill (\xslant*0.222,-0.05*\yslant) circle (\pt) node ('R') {};
       		\fill (\xslant*-0.249,-0.121*\yslant) circle (\pt) node ('S') {};
		\fill (\xslant*-0.19,0.003*\yslant) circle (\pt) node ('T') {};
		\fill (\xslant*-0.14,-0.24*\yslant) circle (\pt) node ('U') {};
		
		\fill (\xslant*0.319,0.009*\yslant) circle (\pt) node ('V') {};
       		\fill (\xslant*-0.28,0.276*\yslant) circle (\pt) node ('W') {};
		\fill (\xslant*0.27,0.4*\yslant) circle (\pt) node ('X') {};
		\fill (\xslant*0.29,-0.21*\yslant) circle (\pt) node ('Y') {};
		
		\fill (\xslant*0.333,-0.22*\yslant) circle (\pt) node ('Z') {};
       		\fill (\xslant*-0.4,-0.01*\yslant) circle (\pt) node ('AA') {};
		\fill (\xslant*-0.17,0.28*\yslant) circle (\pt) node ('BB') {};
		\fill (\xslant*-0.24,-0.25*\yslant) circle (\pt) node ('CC') {};
       
       		{\color{gray} \draw[thick,-] ('L') -- node[above]  {$3$} ('X'); }

        \end{scope}
        
	\renewcommand{\xslant}{1.0}
	\renewcommand{\yslant}{1.0}
 
        \begin{scope}[shift={(5.55cm,-5cm)}]
 
		\fill (\xslant*0,0*\yslant) circle (\pt) node ('A') {};
		
		\fill (\xslant*0.079,0.15*\yslant) circle (\pt) node ('B') {};
       		\fill (\xslant*-0.12,0.04449*\yslant) circle (\pt) node ('C') {};
		\fill (\xslant*0.06,0.19*\yslant) circle (\pt) node ('D') {};
		\fill (\xslant*0.3,-0.11*\yslant) circle (\pt) node ('E') {};
		
		\fill (\xslant*0.204,0.18*\yslant) circle (\pt) node ('F') {};
       		\fill (\xslant*-0.05,0.003*\yslant) circle (\pt) node ('G') {};
		\fill (\xslant*0.1,0.23*\yslant) circle (\pt) node ('H') {};
		\fill (\xslant*0.09,-0.221*\yslant) circle (\pt) node ('I') {};

		\fill (\xslant*0.3,0*\yslant) circle (\pt) node ('J') {};
       		\fill (\xslant*-0.34,0.03*\yslant) circle (\pt) node ('K') {};
		\fill (\xslant*0.007,0.31*\yslant) circle (\pt) node ('L') {};
		\fill (\xslant*0.09,-0.39*\yslant) circle (\pt) node ('M1') {};

		\fill (\xslant*0.28,0.09*\yslant) circle (\pt) node ('N') {};
       		\fill (\xslant*-0.0178,0.13*\yslant) circle (\pt) node ('O') {};
		\fill (\xslant*0.113,0.269*\yslant) circle (\pt) node ('P') {};
		\fill (\xslant*0.164,-0.205*\yslant) circle (\pt) node ('Q') {};
		
		\fill (\xslant*0.222,-0.05*\yslant) circle (\pt) node ('R') {};
       		\fill (\xslant*-0.249,-0.121*\yslant) circle (\pt) node ('S') {};
		\fill (\xslant*-0.19,0.003*\yslant) circle (\pt) node ('T') {};
		\fill (\xslant*-0.14,-0.24*\yslant) circle (\pt) node ('U') {};
		
		\fill (\xslant*0.319,0.009*\yslant) circle (\pt) node ('V') {};
       		\fill (\xslant*-0.28,0.276*\yslant) circle (\pt) node ('W') {};
		\fill (\xslant*0.29,-0.21*\yslant) circle (\pt) node ('Y') {};
		
		\fill (\xslant*0.333,-0.22*\yslant) circle (\pt) node ('Z') {};
       		\fill (\xslant*-0.4,-0.01*\yslant) circle (\pt) node ('AA') {};
		\fill (\xslant*-0.17,0.28*\yslant) circle (\pt) node ('BB') {};
		\fill (\xslant*-0.24,-0.25*\yslant) circle (\pt) node ('CC') {};
       
        \end{scope}
      
      	\renewcommand{\xslant}{0.8}
	\renewcommand{\yslant}{1.3}
        
        \begin{scope}[shift={(6.5cm,-5cm)}]
 
		\fill (\xslant*0,0*\yslant) circle (\pt) node ('A') {};
		
		\fill (\xslant*0.079,0.15*\yslant) circle (\pt) node ('B') {};
       		\fill (\xslant*-0.12,0.04449*\yslant) circle (\pt) node ('C') {};
		\fill (\xslant*0.06,0.19*\yslant) circle (\pt) node ('D') {};
		\fill (\xslant*0.3,-0.11*\yslant) circle (\pt) node ('E') {};
		
		\fill (\xslant*0.204,0.18*\yslant) circle (\pt) node ('F') {};
       		\fill (\xslant*-0.05,0.003*\yslant) circle (\pt) node ('G') {};
		\fill (\xslant*0.1,0.23*\yslant) circle (\pt) node ('H') {};
		\fill (\xslant*0.09,-0.221*\yslant) circle (\pt) node ('I') {};

		\fill (\xslant*0.3,0*\yslant) circle (\pt) node ('J') {};
       		\fill (\xslant*-0.34,0.03*\yslant) circle (\pt) node ('K') {};
		\fill (\xslant*0.007,0.31*\yslant) circle (\pt) node ('L') {};
		\fill (\xslant*0.09,-0.39*\yslant) circle (\pt) node ('M2') {};

		\fill (\xslant*0.28,0.09*\yslant) circle (\pt) node ('N') {};
       		\fill (\xslant*-0.0178,0.13*\yslant) circle (\pt) node ('O') {};
		\fill (\xslant*0.113,0.269*\yslant) circle (\pt) node ('P') {};
		\fill (\xslant*0.164,-0.205*\yslant) circle (\pt) node ('Q') {};
		
		\fill (\xslant*0.222,-0.05*\yslant) circle (\pt) node ('R') {};
       		\fill (\xslant*-0.249,-0.121*\yslant) circle (\pt) node ('S') {};
		\fill (\xslant*-0.19,0.003*\yslant) circle (\pt) node ('T') {};
		\fill (\xslant*-0.14,-0.24*\yslant) circle (\pt) node ('U') {};
		
		\fill (\xslant*0.319,0.009*\yslant) circle (\pt) node ('V') {};
       		\fill (\xslant*-0.28,0.276*\yslant) circle (\pt) node ('W') {};
		\fill (\xslant*0.26,0.35*\yslant) circle (\pt) node ('X') {};
		\fill (\xslant*0.29,-0.21*\yslant) circle (\pt) node ('Y') {};
		
		\fill (\xslant*0.333,-0.22*\yslant) circle (\pt) node ('Z') {};
       		\fill (\xslant*-0.4,-0.01*\yslant) circle (\pt) node ('AA') {};
		\fill (\xslant*-0.17,0.28*\yslant) circle (\pt) node ('BB') {};
		\fill (\xslant*-0.24,-0.25*\yslant) circle (\pt) node ('CC') {};
       
       		{\color{gray} \draw[thick,-] ('M1') -- node[below] {$2$} ('M2');}

        \end{scope}
\end{tikzpicture}
\caption{Two-dimensional use spaces \citep{Tuggy1993,Zlatev03p447} in two time periods with a target word $w$ undergoing innovative meaning change. Dots represent uses of $w$. {\color{gray} Spatial proximity} of two uses means high relatedness.}
\label{fig:usewithin}
\end{center}
\end{figure}

We frame our interest in lexical semantic change as judging the strength of semantic relatedness across use pairs of a target word $w$ within a specific time period $t_i$.
A high mean proximity value indicates meaning identity or context variance, while a low value indicates polysemy or homonymy, cf. Table \ref{tab:blank}. This strategy is applied independently to two time periods $t_1$ and $t_2$, as illustrated in Figure \ref{fig:usewithin}.
Innovative vs. reductive meaning change can then be measured by decrease vs. increase in the mean relatedness value of $w$ from $t_1$ to $t_2$. To see why this is justified, consider the different semantic constellations of $w$'s use pairs in $t_1$ and $t_2$ in Figure \ref{fig:usewithin}. If $w$ is monosemous in $t_1$ and undergoes innovative meaning change between the two time periods, we expect to find use pairs in the later period $t_2$ combining the old and new meaning which are less related (score: 2) than the use pairs from the earlier period $t_1$ (score: 3).
According to this rationale, the mean relatedness values across $w$'s use pairs should be lower in $t_2$ than in $t_1$. The reverse applies to reductive meaning change.

There are a number of other, more complex semantic constellations. For example, if $w$ not only gains a new meaning, but rapidly loses the old meaning, we cannot necessarily expect the mean relatedness of $w$'s use pairs to be higher in the later than in the earlier time period. In order to cover such cases, we will not only measure the mean relatedness within the \textsc{earlier} and the \textsc{later} groups of use pairs but also in a mixed \textsc{compare} group where each pair consists of a use from the \textsc{earlier} and a use from the \textsc{later} group. By this, old and new meaning are directly compared, and we do not have to rely on the assumption that the old meaning is still present.

By applying the above-described procedure to all target words and sorting them according to their mean relatedness scores, we obtain a ranked list for each of the three groups \textsc{earlier}, \textsc{later} and \textsc{compare}. We then exploit two measures of change:  (i) $\Delta$\textsc{later} measures changes in the degree of mean relatedness of words, and is derived by subtracting a target $w$'s mean in \textsc{earlier} from its mean in \textsc{later}: $\Delta\textsc{later}(w) = Mean_{l}(w)-Mean_{e}(w)$. Positive vs. negative values on this measure indicate innovative vs. reductive meaning change. (ii) \textsc{compare} directly measures the relatedness \textit{between} the \textsc{earlier} and the \textsc{later} group and thus corresponds to $w$'s mean in the \textsc{compare} group: $\textsc{compare}(w) = Mean_{c}(w)$. High vs. low values on \textsc{compare} indicate weak vs. strong change, where the change includes both innovative and reductive meaning changes.

\section{Annotation Study}
\label{section:annotation}

Five native speakers of German were asked to rate 1,320 use pairs on our 4-point scale of relatedness in Table \ref{tab:scale2}. All annotators were students of linguistics. We explicitly chose two annotators with a background in historical linguistics in order to see whether knowledge about historical linguistics has an effect on the annotation. Annotators were not told that the study is related to semantic change.\footnote{The guidelines (adapted from the synchronic study by \citealp{Erk13}) and the experiment data are publicly available at \url{www.ims.uni-stuttgart.de/data/durel/}.}

\paragraph{Target Words.}  The target words were selected by manually checking a corpus for innovative and reductive meaning changes, based on cases of metaphoric, metonymic change and narrowing (innovative) as reported by \citet{Paul02XXI}, and cases of reduction due to homonymy (reductive) as reported by \citet{Osman1971}. The corpus we used is DTA \citep{dta2017}, a freely available diachronic corpus of German. By focusing on a late time period (19th century), we tried to reduce problems coming with historical language data as much as possible. We still normalized special characters to modern orthography.

We included only those words as targets for which we found the change suggested by the literature reflected in the corpus, either weakly or strongly, because an annotation relying on a random selection of words suggested to undergo change is likely to produce a set with very similar and rather low values representing small effects. We thus guaranteed to include both: words for which we expected weak effects as well as words for which we expected strong effects. We ended up with 19 cases of innovation and 9 cases of reduction. Three words, \textit{Anstalt}, \textit{Anstellung} and \textit{Vorwort} represent especially interesting cases and were selected more than once for the test set since they undergo both innovative \textit{and} reductive change between the investigated time periods.

\paragraph{Sampling.} For each target word we randomly sampled 20 use pairs from DTA (searching for the respective lemma and POS) for each of the groups \textsc{earlier} (1750-1800), \textsc{later} (1850-1900) and \textsc{compare}, yielding 60 use pairs per word and 1,320 use pairs for 22 target words in total.

\begin{figure}
  \centering
  \includegraphics[width=1.0\linewidth]{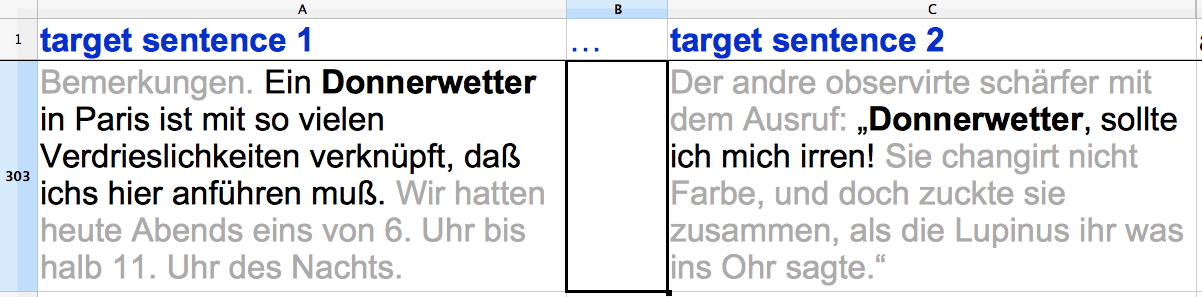}
  \caption{Use pair from annotation table (English adaption).}
\label{fig:anno}  
\end{figure}

A use of a word is defined as the sentence the word occurs in. The annotators were provided these sentences as well as the preceding and the following sentence in the corpus, cf. Figure \ref{fig:anno}. We double-checked that each use of a word was only sampled once within each group. If the total number of uses in the group was less than needed, uses were allowed twice across pairs. Before presenting the use pairs to the annotators in a spreadsheet, uses within pairs were randomized, and pairs from all groups were mixed and randomly ordered.

\captionsetup{belowskip=-40pt}
\captionsetup{belowskip=\globalcaptionsetup}

\section{Results}
\label{sec:results}

\paragraph{Agreement.}
In line with \citet{Erk13} we measure inter-annotator agreement as the average over pairwise Spearman’s $\rho$ correlations (omitting 0-judgments), cf. Table \ref{tab:correlation}. The bottom line provides the agreement of each annotator’s judgments against the average judgment score across the other annotators. The range of correlation coefficients is between 0.57 and 0.68, with an average correlation of 0.66. All the pairs are highly significantly correlated ($p<0.01$).

\captionsetup{belowskip=-20pt}
\begin{table}[htbp]
\begin{center}

\begin{tabular}{llllll}
\hline
  & 1 & 2 & 3 & 4 & 5 \\\hline  
1 & & 0.59 & 0.63 & 0.67 & 0.66 \\  
2 & &  & 0.57 & 0.64 & 0.65 \\  
3 & &  &  & 0.64 & 0.62 \\  
4 & &  &  &  & 0.68 \\  
 &  & &  &  &  \\  
\textit{avg} & 0.71 & 0.68 & 0.68 & 0.75 & 0.74 \\ 
\end{tabular}
\caption{Correlation matrix for pairwise correlation agreement; \textit{avg} refers to agreement of the annotator in the column against the average across the other annotators.}
\label{tab:correlation}
\end{center}
\end{table}
\captionsetup{belowskip=\globalcaptionsetup}

The annotators with historical background are annotators 4 and 5, who show the highest pairwise agreement and also the highest agreement with the average of the other annotators. This indicates that historical knowledge makes a positive difference when annotating DURel. Yet, the agreements of the non-expert annotators only deviate slightly.

Overall, our correlations are comparable and even moderately higher than the ones found in \citet{Erk13}, who report average correlation scores between 0.55 and 0.62. This difference is remarkable, given that annotators had to judge historical data. Note, however, that the studies are not exactly comparable, as \citeauthor{Erk13} used a more fine-grained 5-point scale, and we presumably excluded a larger number of 0-judgments.

\begin{figure}
  \centering
  \includegraphics[width=1.0\linewidth]{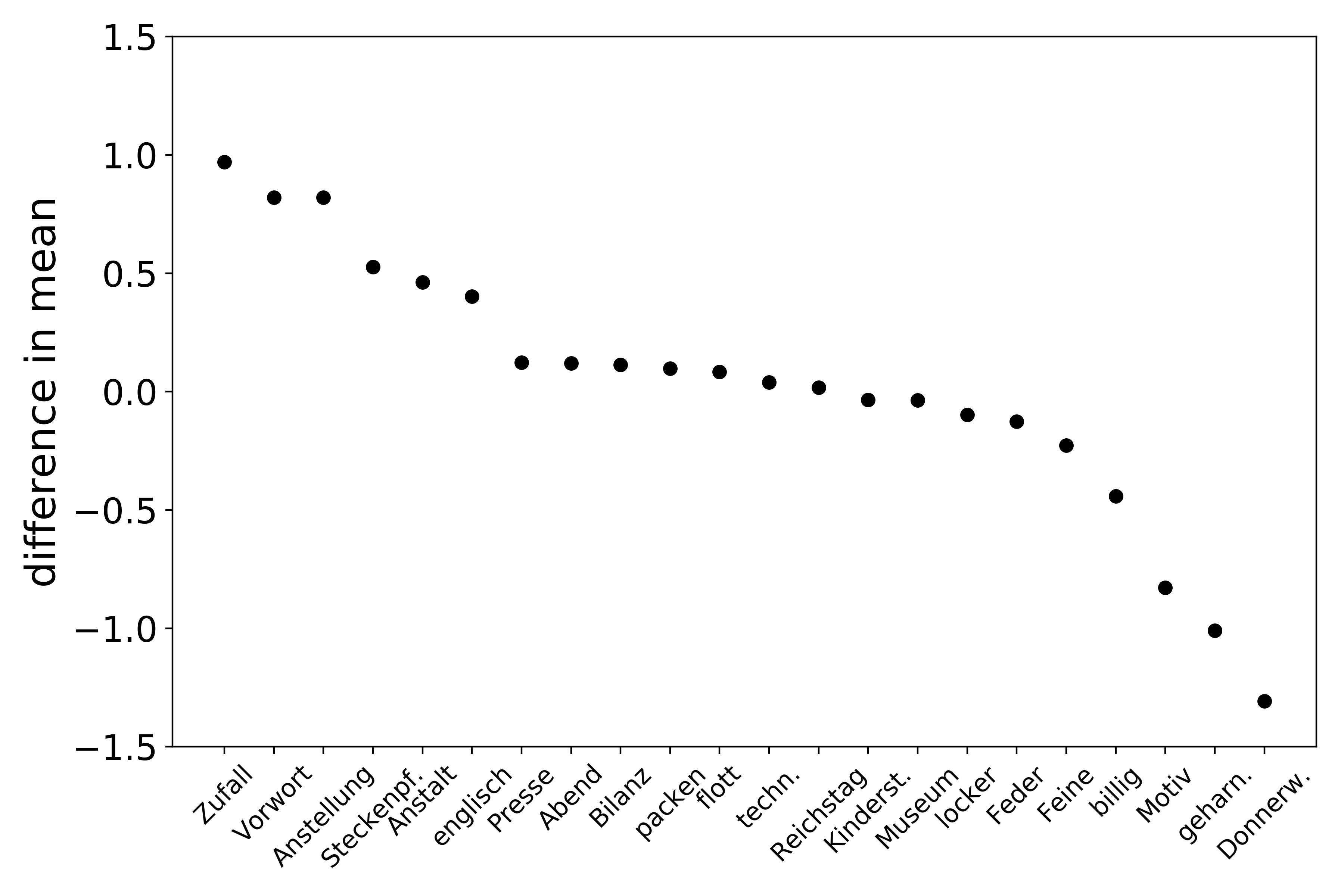}
  \caption{$\Delta$\textsc{later}: Rank of target words according to increase in mean usage relatedness from \textsc{earlier} to \textsc{later}.}
\label{fig:ranks}  
\vspace{+3mm}
\end{figure}

\paragraph{Qualitative Analysis.}
Figure \ref{fig:ranks} shows the target words ranked according to their values on $\Delta$\textsc{later}. We can clearly identify three groups: words with values $>$0, $<$0, and a majority with values $\approx$0 difference in mean between the earlier and the later time period. The three topmost words have previously been classified as reductive, the three lowermost words as innovative meaning changes.

\begin{figure}
 \begin{subfigure}{.5\linewidth}
  \centering
   \includegraphics[width=1.0\linewidth]{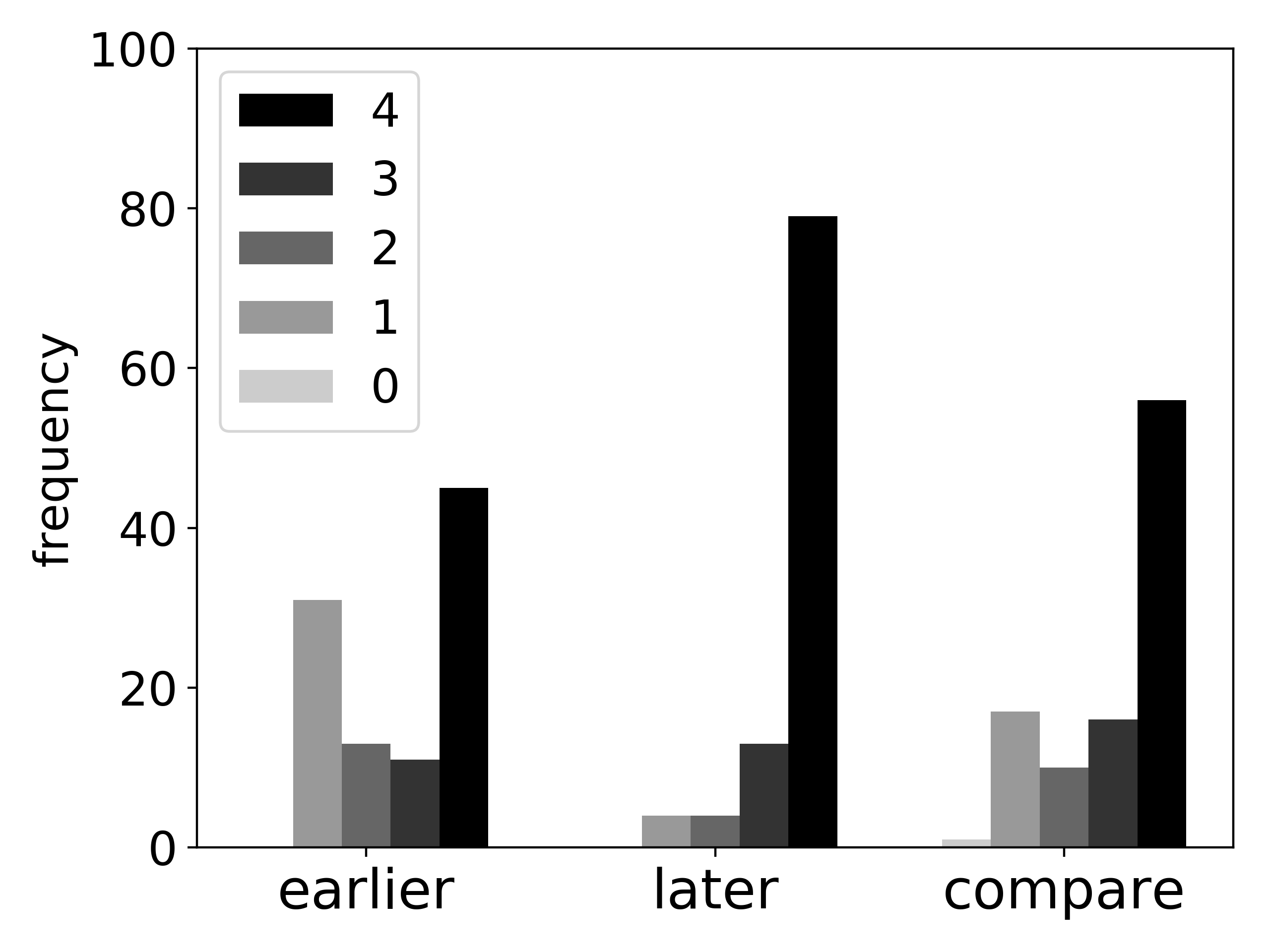} 
  \caption*{\textit{Zufall} (reductive)}
 \end{subfigure}%
\begin{subfigure}{.5\linewidth}
  \centering
  \includegraphics[width=1.0\linewidth]{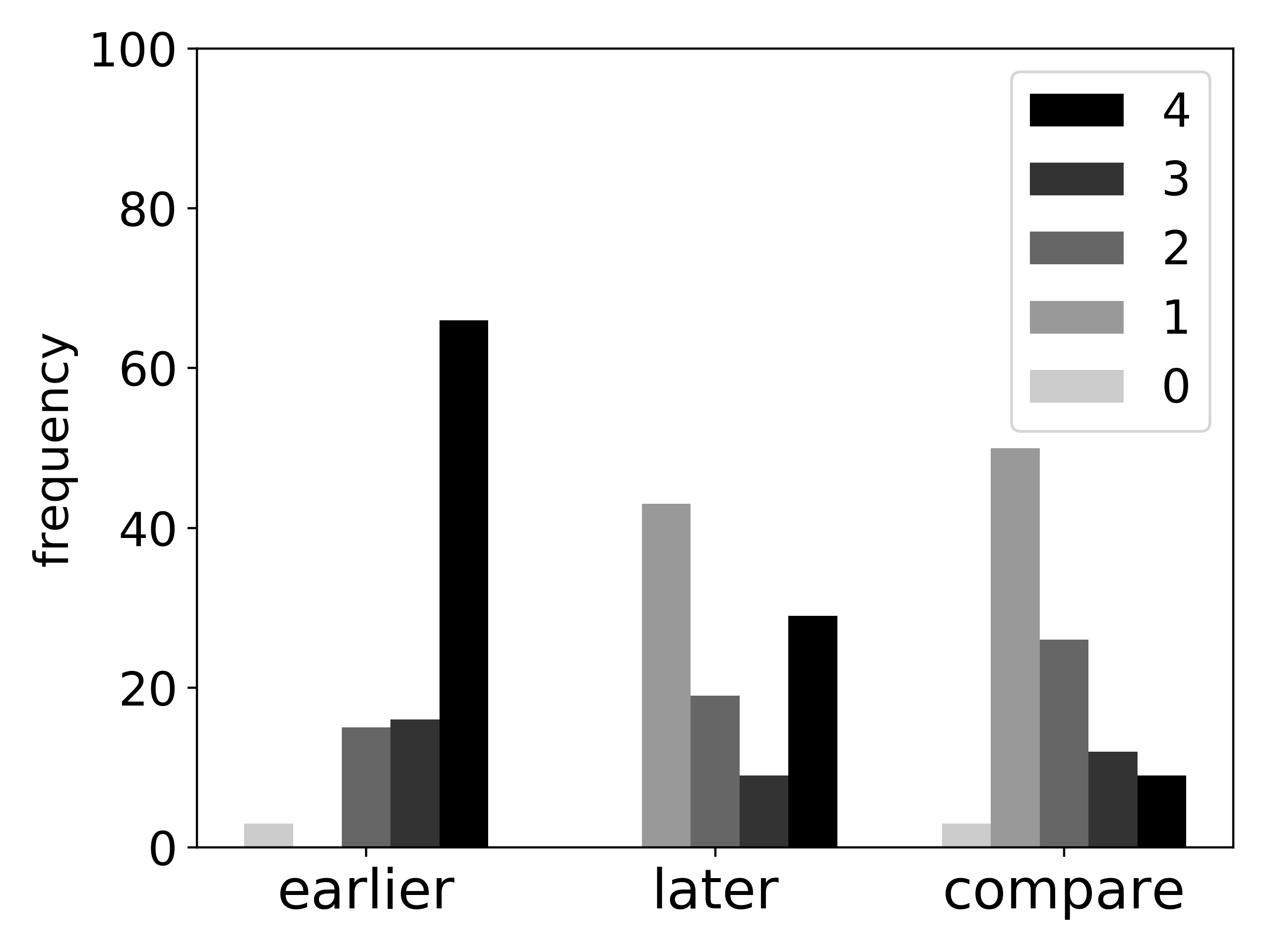}
  \caption*{\textit{Donnerwetter} (innovative)}
\end{subfigure}%
\captionsetup{belowskip=-30pt}
\caption{Plots of judgment freq. for target words per group.}
\label{fig:each_plots}
\end{figure}

Figure \ref{fig:each_plots} compares the distributions across relatedness scores for our two example words \textit{Zufall} and \textit{Donnerwetter} from above. In \textsc{earlier}, \textit{Zufall}'s ratings (i.e., the number of times a specific rating 0--4 was provided) vary much more than in \textsc{later} where it has a high number of 4-judgments. The contrary is the case for \textit{Donnerwetter}. In addition, we find a clear difference between the two words in the \textsc{compare} group, because \textit{Donnerwetter} is used in a variety of new figurative ways in \textsc{later}, while \textit{Zufall}, besides losing the meaning `seizure', retains the prevalent meaning `coincidence' in both time periods. 

Upon closer inspection, the words deviating most from our predictions show either that the change is already present before 1800 (e.g., \textit{Steckenpferd}, `toy $>$ toy; hobby'), that the new meaning has a very low prevalence (e.g., \textit{Museum}, `study room; arts collection $>$ arts collection'), or that there are additional, previously not identified uses in the later time period (e.g., \textit{Feine}, `fineness; grandeur $>$ grandeur'). The mean value for reduction is 0.39, while it is -0.18 for innovation.

Overall, these findings confirm our predictions and validate $\Delta$\textsc{later} as a measure of lexical semantic change. 
The case of \textit{Presse}, `printing press $>$ printing press; print product/producer', however, shows its shortcomings: $\Delta$\textsc{later} wrongly predicts no change for \textit{Presse}, although it is clearly present, because the new meaning has a very high prevalence. $\Delta$\textsc{later} cannot capture such cases, while \textsc{compare} can: it predicts strong change for \textit{Presse}.

Since \textsc{compare} measures the degree of change rather than distinguishing between types of change, the highest values in its ranked list refer to cases with values$\approx$0 in $\Delta$\textsc{later}, and the lowest values refer to cases with extreme values of $\Delta$\textsc{later}. A special case is \textit{Feder} `bird feather $>$ bird feather; steel clip', which reveals the need for normalization of the \textsc{compare}-measure: the word is highly polysemous and has approximately the same distribution in every group, because the new meaning `steel clip' has a very low prevalence. For $\Delta$\textsc{later} this correctly leads to a 0-prediction. In contrast, \textsc{compare} predicts strong change, because due to polysemy there is a high probability to sample distantly related use pairs in the \textsc{compare} group.

\captionsetup{belowskip=-0pt}
\captionsetup{belowskip=\globalcaptionsetup}

\captionsetup{belowskip=-5pt}
\captionsetup{belowskip=\globalcaptionsetup}

\paragraph{Discussion.} While our measures enable us to predict various semantic change constellations, we also demonstrated that they collapse in certain semantic constellations: $\Delta$\textsc{later} is accurate when used for simple semantic constellations (i.e., only one reductive or innovative meaning change), where the old meaning roughly maintains its prevalence, thus making  $\Delta$\textsc{later} be prone to corpus effects such as changes in text genre. For an optimal application of this measure we therefore recommend (i) to choose directly adjacent and short time periods for annotation, as the number of changes increases with the length of the time period, and (ii) to use a well-balanced  corpus for the annotation, ideally across all periods.

Unlike $\Delta$\textsc{later}, \textsc{compare} has the advantage to capture multiple changes over time, but it confuses polysemy and meaning change. In future work, we aim to solve this issue by normalizing \textsc{compare} with a measure of polysemy: For any target word $w$ the values from the \textsc{earlier} group determine its degree of polysemy in the earlier time period. Hence, the normalized $\Delta\textsc{compare}(w) = Mean_{c}(w)-Mean_{e}(w)$ intuitively measures how much the values in the \textsc{compare} group differ from what we would already expect from $w$'s early polysemy, so it predicts no change in the case of a stable polysemous word, and it predicts change if the word gains or loses a meaning.

\vspace{+3mm}
\section{Conclusion}
\label{sec:conclusion}

This paper presented a general framework DURel for language-independent annotation of Diachronic Usage Relatedness, in order to develop test sets for lexical semantic change. In addition to a strong theoretical basis, DURel shows empirical validity in our annotation study with high inter-annotator agreement. It relies on an intuitive notion of semantic relatedness and needs no definition of word senses. 

Furthermore, we proposed two measures of lexical semantic change that predict various semantic change constellations. While one measure successfully distinguishes between innovative and reductive meaning change, we also demonstrated the need to refine and normalize the measures in order to capture more variants of constellations regarding the interplay of polysemy and meaning reduction/innovation.

The annotated test set for German is publicly available and can be used to compare computational models of semantic change, and more generally to evaluate models of lexical variation in corpora across times, domains, etc. Further test sets across languages can be obtained by applying DURel to the respective language uses.

\vspace{+5mm}
\section*{Acknowledgments}

The research was supported by the DFG Collaborative Research Centre SFB 732. We thank Katrin Erk, Diana McCarthy and Susan Brown Windisch for providing their expertise, experience and data; Eleonore Brandner, Jun Chen, Fabian Bross and Diego Frassinelli for helpful discussions and comments; Delia Alf, Pavlos Musenidis, Cornelia van Scherpenberg and Jennifer Schild for help with the annotations; and the three reviewers for their constructive criticism.

\vspace{+3mm}
\bibliographystyle{acl_natbib}
\bibliography{/Users/admin/Documents/workspace-python/backup/literature/Bibliography-general.bib}

\begin{thebibliography}{}
\expandafter\ifx\csname natexlab\endcsname\relax\def\natexlab#1{#1}\fi

\bibitem[{Bamman and Crane(2011)}]{Bamman11p1}
D.~Bamman and G.~Crane. 2011.
\newblock Measuring historical word sense variation.
\newblock In {\em {Proceedings of the 11th Annual International ACM/IEEE Joint
  Conference on Digital Libraries}\/}. ACM, New York, NY, USA, pages 1--10.

\bibitem[{Blank(1997)}]{Blank97XVI}
A.~Blank. 1997.
\newblock {\em {Prinzipien des lexikalischen Bedeutungswandels am Beispiel der
  romanischen Sprachen}\/}.
\newblock Niemeyer, T{\"u}bingen.

\bibitem[{Brown(2008)}]{Brown2008}
S.~W. Brown. 2008.
\newblock Choosing sense distinctions for {WSD}: {Psycholinguistic} evidence.
\newblock In {\em {Proceedings of the 46th Annual Meeting of the Association
  for Computational Linguistics on Human Language Technologies: Short
  Papers}\/}. Stroudsburg, PA, USA, pages 249--252.

\bibitem[{Cook et~al.(2014)Cook, Lau, McCarthy, and Baldwin}]{Cook14p1624}
P.~Cook, J.~H. Lau, D.~McCarthy, and T.~Baldwin. 2014.
\newblock Novel word-sense identification.
\newblock In {\em {25th International Conference on Computational Linguistics,
  Proceedings of the Conference: Technical Papers}\/}.  Dublin,
  Ireland, pages 1624--1635.

\bibitem[{{Deutsches Textarchiv}(2017)}]{dta2017}
{Deutsches Textarchiv}. 2017.
\newblock \href{http://www.deutschestextarchiv.de/}{{Grundlage für ein
  Referenzkorpus der neuhochdeutschen Sprache. Herausgegeben von der
  Berlin-Brandenburgischen Akademie der Wissenschaften}}.
\newblock
  \href{http://www.deutschestextarchiv.de/}{http://www.deutschestextarchiv.de/}.

\bibitem[{Dubossarsky et~al.(2017)Dubossarsky, Weinshall, and
  Grossman}]{dubossarsky2017}
Haim Dubossarsky, Daphna Weinshall, and Eitan Grossman. 2017.
\newblock Outta control: Laws of semantic change and inherent biases in word
  representation models.
\newblock In {\em {Proceedings of the 2017 Conference on Empirical Methods in
  Natural Language Processing}\/}. Copenhagen, Denmark, pages 1147--1156.

\bibitem[{Erk et~al.(2009)Erk, McCarthy, and Gaylord}]{Erk09investigationson}
K.~Erk, D.~McCarthy, and N.~Gaylord. 2009.
\newblock Investigations on word senses and word usages.
\newblock In {\em Proceedings of the Joint Conference of the 47th Annual
  Meeting of the ACL and the 4th International Joint Conference on Natural
  Language Processing of the AFNLP: Volume 1\/}. Stroudsburg, PA,
  USA, pages 10--18.

\bibitem[{Erk et~al.(2013)Erk, McCarthy, and Gaylord}]{Erk13}
K.~Erk, D.~McCarthy, and N.~Gaylord. 2013.
\newblock Measuring word meaning in context.
\newblock {\em Computational Linguistics\/} 39(3):511--554.

\bibitem[{Frermann and Lapata(2016)}]{Frermann:2016}
L.~Frermann and M.~Lapata. 2016.
\newblock A {Bayesian} model of diachronic meaning change.
\newblock {\em Transactions of the Association for Computational Linguistics\/}
  4:31--45.

\bibitem[{Gulordava and Baroni(2011)}]{Gulordava11}
K.~Gulordava and M.~Baroni. 2011.
\newblock A distributional similarity approach to the detection of semantic
  change in the {Google Books Ngram} corpus.
\newblock In {\em {Proceedings of the Workshop on Geometrical Models of Natural
  Language Semantics}\/}. Stroudsburg, PA, USA, pages 67--71.

\bibitem[{Hamilton et~al.(2016)Hamilton, Leskovec, and Jurafsky}]{Hamilton16}
W.~L. Hamilton, J.~Leskovec, and D.~Jurafsky. 2016.
\newblock Cultural shift or linguistic drift? {Comparing} two computational
  measures of semantic change.
\newblock In {\em {Proceedings of the 2016 Conference on Empirical Methods in
  Natural Language Processing}\/}. Austin, Texas, pages 2116--2121.

\bibitem[{Ide and Pustejovsky(2017)}]{IdePustejovsky2017}
N.~Ide and J.~Pustejovsky, editors. 2017.
\newblock {\em Handbook of Linguistic Annotation\/}.
\newblock Springer, Dordrecht.

\bibitem[{Lau et~al.(2012)Lau, Cook, McCarthy, Newman, and Baldwin}]{Lau12p591}
J.~H. Lau, P.~Cook, D.~McCarthy, D.~Newman, and T.~Baldwin. 2012.
\newblock Word sense induction for novel sense detection.
\newblock In {\em {Proceedings of the 13th Conference of the European Chapter
  of the Association for Computational Linguistics}\/}. Stroudsburg, PA, USA,
  pages 591--601.

\bibitem[{Osman(1971)}]{Osman1971}
N.~Osman. 1971.
\newblock {\em {Kleines Lexikon untergegangener Wörter: Wortuntergang seit dem
  Ende des 18. Jahrhunderts}\/}.
\newblock Beck, München.

\bibitem[{Paul(2002)}]{Paul02XXI}
H.~Paul. 2002.
\newblock {\em {Deutsches W{\"o}rterbuch: Bedeutungsgeschichte und Aufbau
  unseres Wortschatzes}\/}.
\newblock Niemeyer, T{\"u}bingen, 10. edition.

\bibitem[{Rosch and Mervis(1975)}]{Rosch1975}
E.~Rosch and C.B. Mervis. 1975.
\newblock Family resemblances: {Studies} in the internal structure of
  categories.
\newblock {\em Cognitive Psychology\/} 7:573--605.

\bibitem[{Schlechtweg et~al.(2017)Schlechtweg, Eckmann, Santus, {Schulte im
  Walde}, and Hole}]{schlechtweg-EtAl:2017:CoNLL}
D.~Schlechtweg, S.~Eckmann, E.~Santus, S.~{Schulte im Walde}, and D.~Hole.
  2017.
\newblock German in flux: {Detecting} metaphoric change via word entropy.
\newblock In {\em {Proceedings of the 21st Conference on Computational Natural
  Language Learning}\/}. Vancouver, Canada, pages 354--367.

\bibitem[{Soares~da Silva(1992)}]{SoaresdaSilva1992}
A.~Soares~da Silva. 1992.
\newblock Homon{\'i}mia e polissemia: An{\'a}lise s{\'e}mica e teoria do
  campol{\'e}xico.
\newblock In {\em Actas do XIX Congreso Internacional de Ling\"{u}{\'i}stica e
  Filolox{\'i}a Rom{\'a}nicas\/}. Fundaci{\'o}n Pedro Barri{\'e} de la Maza, La
  Coru{\~n}a, volume~2 of {\em Lexicolox{\'i}a e Metalexicograf{\'i}a\/}, pages
  257--287.

\bibitem[{Tuggy(1993)}]{Tuggy1993}
David Tuggy. 1993.
\newblock Ambiguity, polysemy, and vagueness.
\newblock {\em Cognitive Linguistics\/} 4(3):273--290.

\bibitem[{Zlatev(2003)}]{Zlatev03p447}
J.~Zlatev. 2003.
\newblock {\em {Polysemy or generality? Mu}\/}, Mouton de Gruyter, volume~23 of
  {\em Cognitive Approaches to Lexical Semantics\/}, pages 447--494.

\end{thebibliography}

\end{document}